\documentclass[runningheads]{llncs}

\usepackage{eccv}

\usepackage{eccvabbrv}

\usepackage{graphicx}
\usepackage{booktabs}

\usepackage[accsupp]{axessibility}  
\usepackage{booktabs}
\usepackage{makecell}
\usepackage{array}
\usepackage{tabularx}

\usepackage{hyperref}

\usepackage{orcidlink}

\begin{document}

\title{Human-like Object Grouping in Self-supervised Vision Transformers}   


\titlerunning{Human-like Grouping in Transformers}

\author{Hossein Adeli\inst{1}\orcidlink{0000-0002-1132-1091} \and
Seoyoung Ahn\inst{2}\and
Andrew Luo\inst{3} \and
Mengmi Zhang\inst{4} \and
Nikolaus Kriegeskorte \inst{1} \and
Gregory Zelinsky \inst{5}}


\institute{Zuckerman Mind Brain Behavior Institute, Columbia University, New York \and
Department of Social Science and AI, Hankuk University of Foreign Studies, Seoul \and
University of Hong Kong, Hong Kong \and
Nanyang Technological University, Singapore \and
Stony Brook University, New York \\
\email{ha2366@columbia.edu}}


\authorrunning{Adeli et al.}


\maketitle

\begin{abstract}

Vision foundation models trained with self-supervised objectives achieve strong performance across diverse tasks and exhibit emergent object segmentation properties. However, their alignment with human object perception remains poorly understood. Here, we introduce a behavioral benchmark in which participants make same/different object judgments for dot pairs on naturalistic scenes, scaling up a classical psychophysics paradigm to over 1000 trials. We test a diverse set of vision models using a simple readout from their representations to predict subjects' reaction times. We observe a steady improvement across model generations, with both architecture and training objective contributing to alignment, and transformer-based models trained with the DINO self-supervised objective showing the strongest performance. To investigate the source of this improvement, we propose a metric to quantify the object-centric component of representations by measuring patch similarity within and between objects. Across models, stronger object-centric structure predicts human segmentation behavior more accurately. We further show that matching the Gram matrix of supervised transformer models, capturing similarity structure across image patches, with that of a self-supervised model through distillation improves their alignment with human behavior, converging with the prior finding that Gram anchoring improves DINOv3's feature quality. Together, these results demonstrate that self-supervised vision models capture object structure in a behaviorally human-like manner, and that Gram matrix structure plays a role in driving perceptual alignment. \href{https://github.com/Hosseinadeli/human_like_grouping_Transformers}{Code and data are available here.}


\end{abstract}

\begin{figure}[tb]
  \begin{center}
    \centerline{\includegraphics[width=0.9\columnwidth]{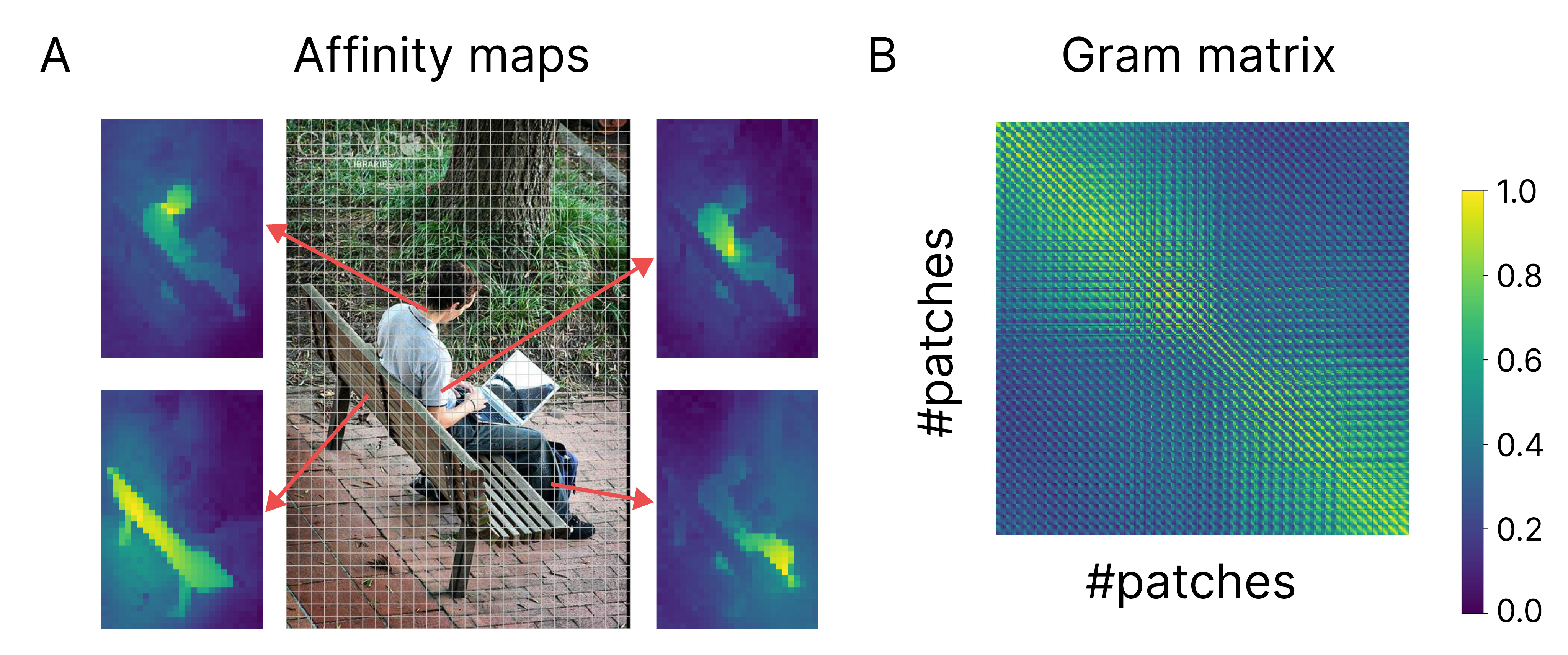}}
    \caption{
      \textbf{A)} Example affinity maps for a few image patches, generated using feature tokens from the DINOv3 ViT-B/16 model. For a given patch, the cosine similarity between its feature vector and the feature vectors of all other patches is shown as an affinity map, with brighter values indicating stronger similarity. Patches belonging to the same object tend to have the highest affinity, reflecting object-centric structure in the representations. \textbf{B)} The full Gram matrix for the same image, showing pairwise feature similarities between all patches. The block-like structure visible along the diagonal reflects clusters of patches with high mutual similarity, corresponding to distinct objects in the scene.
    }
    \label{fig:affinity_maps}
  \end{center}
\end{figure}

\section{Introduction}
\label{sec:intro}

Grouping visual input into coherent whole objects \cite{treisman1996binding, roelfsema2023solving, peters2021capturing} is a fundamental problem for the brain and AI systems. The human visual system achieves this by relying on different signals ranging from part-whole relationships  and Gestalt processes \cite{adeli2023brain, dedieu2021learning, george2017generative, ahn2024attentive, wang2023object, han2024flow, wang2024unsupervised, biswas2026natural, kim2019disentangling, toosi2025generative, chen2022unsupervised}, to prior semantic knowledge of object categories \cite{greff2020binding, vecera2000toward, wagemans2012century, gilbert_top-down_2013, papale2025modulatory}. This visual integration process has been proposed to be mediated by `association fields' \cite{field1993contour}, long-range lateral connections in retinotopic visual cortical areas linking distant points. The effective connectivity between units in these maps has been shown to depend on the similarity between locally represented features (e.g., orientation). Models of object perception have focused on implementing these cues and studying grouping as a gradual activation of lateral connections \cite{roelfsema2023solving, jeurissen2016serial}. However, most behavioral and modeling experiments in this domain have focused on Gestalt cues and objects with clear boundaries \cite{jeurissen2016serial}, and not in natural contexts where objects are less spatially separated or contours lack clear definition.

Recent advances in vision foundation models show that similar object grouping properties can emerge from large-scale self-supervised training of transformer models \cite{caron2021emerging, oquab2023dinov2, simeoni2025dinov3, he2022masked}. The self-attention mechanism in these models forms dynamic contextual connections between different locations of the visual input, similar to association fields forming in retinotopic maps of the visual cortex. Indeed, recent large-scale model comparisons show that contour integration also emerges in these models as training data and model scale increase, with the strongest models beginning to approach human-level integration performance \cite{lonnqvist2025contour}. Contour integration, while a hallmark of human object perception, still puts the focus on low-level Gestalt cues, and it remains to be tested whether these findings extend to naturalistic settings \cite{coen2026we}.

In this work, we address this gap and test whether the object grouping properties of vision foundation models are aligned with human object perception. We first collected human data on a well-controlled behavioral experiment to probe how people group complex objects in naturalistic scenes. Subjects responded to whether two dots placed on a scene were on the same object or two different objects. Their reaction time on this task reveals how they perceive objects and the nuances of how easy or hard it is to group them. We then test whether the patch representations from the models can predict human object grouping behavior. Higher performance on this task indicates that the model representations capture human-like object structure. We then quantify the object-centric component in the patch-level representations of these models as a way to understand what drives this alignment. Figure~\ref{fig:affinity_maps}A shows the affinity maps for a few patches using token features from the DINOv3 model \cite{simeoni2025dinov3}. Each map is created by computing the cosine similarity between the token representation of a patch and all other patches in the image (with higher values shown brighter). Notably, patches that are more similar tend to lie on the same object, suggesting that object identity is reflected in patch-level representations of these models. Our main contributions are as follows.

\begin{itemize}
\item We introduce a large-scale behavioral benchmark in which human participants make same/different object judgments for dot pairs on naturalistic scenes, scaling up a classical psychophysics paradigm to a diverse set of natural images.
\item We propose a novel object-centric metric based on ROC analysis of patch-level affinity maps, which quantifies the degree to which a model's representations reflect object boundaries without requiring object-level supervision.
\item We evaluate a diverse set of vision models and show that self-supervised Transformer models trained with the DINO objective achieve the strongest alignment with human object grouping behavior, with training objective contributing more strongly than architecture in our model set, though these factors are not fully decoupled.
\item We demonstrate that object-centric structure in patch representations, as measured by our proposed metric, is strongly predictive of behavioral alignment across models and training objectives.
\item We show that fine-tuning supervised models to match the Gram matrix structure of a self-supervised model improves their object-centricity and behavioral alignment, converging with independent evidence that Gram anchoring is a key mechanism underlying the dense feature quality of DINOv3.
\end{itemize}

\section{Related Work}

\textbf{Self-supervised Vision Transformers. }
Application of Transformers to vision has been extremely successful, with these models outperforming convolutional neural networks (CNNs) on object recognition and other tasks \cite{dosovitskiy2020image}. The self-attention weights in supervised vision Transformers have been shown to perform some perceptual grouping \cite{lee2022understanding, mehrani2023self, chen2025vision}. More recently, studies have explored training these models with self-supervised objectives, yielding intriguing object-centric properties that are less prominent in models trained for classification. When trained with a self-distillation loss (DINO \cite{caron2021emerging}, DINOv2 \cite{oquab2023dinov2}, and DINOv3 \cite{simeoni2025dinov3}), the attention maps contain explicit information about the semantic segmentation of foreground objects and their parts, reflecting that these models can capture object-centric representations without labels. These models have also been shown to achieve state-of-the-art performance in predicting brain activity in visual areas in response to complex scenes \cite{adeli2025transformer, hwang2025silico}, further pointing to their plausibility as models of human-like visual processing. In masked autoencoding (MAE \cite{he2022masked}), the input image is heavily occluded and the model is trained to reconstruct the full image from a small number of visible patches. Minimizing the reconstruction loss enables the model to learn object-centric features that yield strong performance on downstream tasks.


\noindent\textbf{Self-supervised Transformers for object and part discovery. }
There have been recent attempts to investigate the extent to which self-supervised Transformers can learn high-level characteristics of a scene. These studies involve computing feature similarity among all tokens and examining their correspondence with high-level concepts such as objects and parts. LOST \cite{simeoni2021localizing} and TokenCut \cite{wang2022self} use the similarity graph to perform unsupervised object discovery, showing success when there is one salient object in the scene. Other work \cite{amir2021deep} has used feature similarity to perform co-segmentation of object parts. These results collectively corroborate that vision Transformers trained with a self-supervised objective begin to represent object-centric information, meaning the patches with the highest affinity to a given patch tend to lie on the same object (Fig.~\ref{fig:affinity_maps}A). Consistent with this, recent work using "visual anagrams" to probe configural shape processing shows that self-supervised Transformers exhibit strong sensitivity to global part configuration, relying on long-range interactions and showing a transition from local to global coding across layers \cite{doshi2025visual}. This suggests that these models do not merely rely on local texture cues, but can integrate spatially distributed part information into more holistic object representations \cite{li2025does, li2025local}.

\begin{figure}[tb]
  \centering
  \includegraphics[height=7cm]{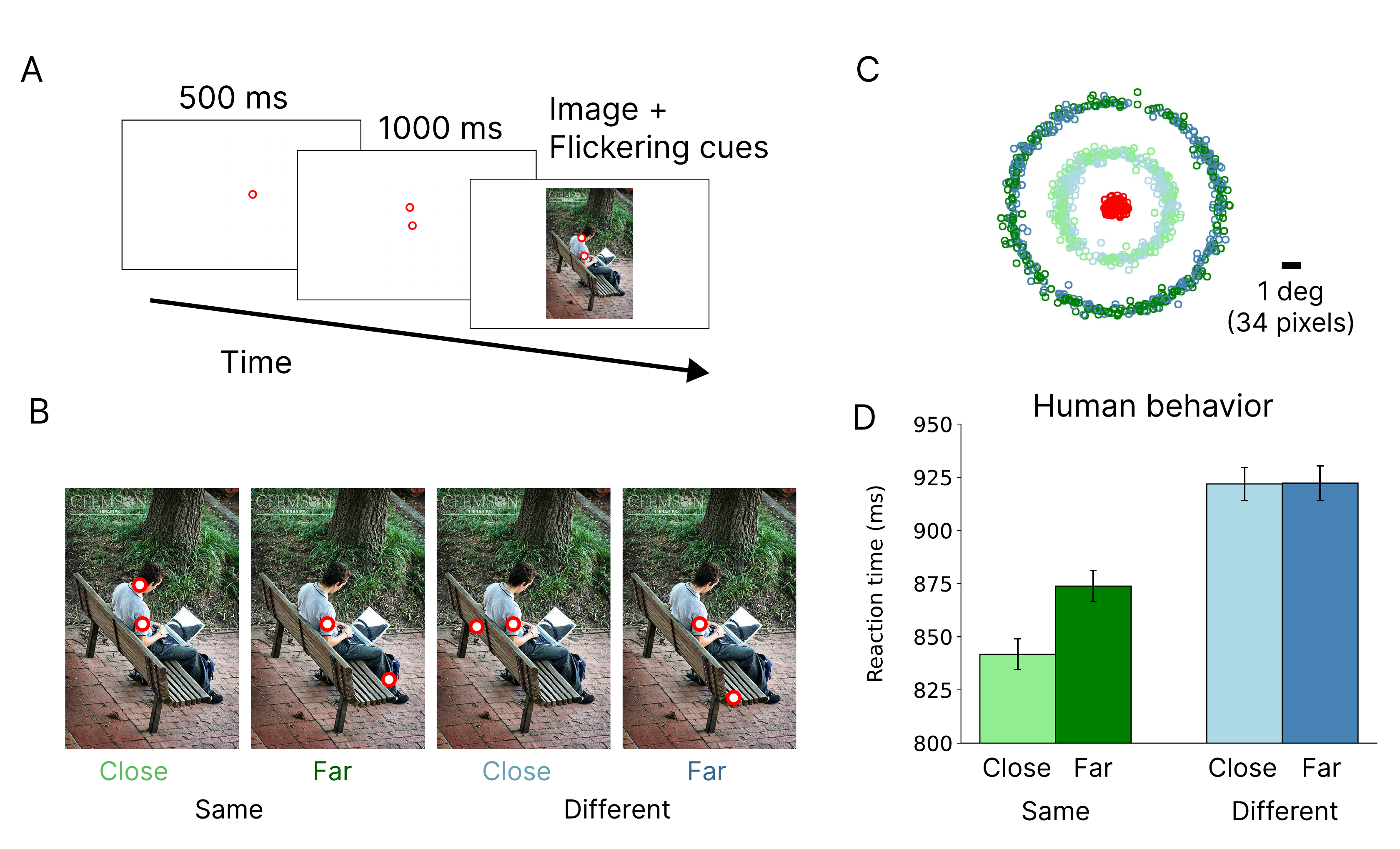} 
  \caption{
    \textbf{A)} Behavioral procedure. Participants maintain fixation on a center dot during the trial. A second dot appears following the center dot and remains visible for 1000 ms, after which the scene appears and the dots begin flickering to ensure their visibility. Subjects are instructed to respond whether the two dots are on the same or two different objects as quickly as possible without sacrificing accuracy. \textbf{B)} Sample trials from all four experimental conditions, coded by different colors. \textbf{C)} Placement of dots across all conditions and trials. \textbf{D)} Mean reaction time for correct trials by condition, with SEM (Standard Error of the Mean) error bars.
}
  \label{fig:beh_exp}
\end{figure}

\section{Behavioral Experiment}

We use a "two-dot" paradigm (Fig.~\ref{fig:beh_exp}) to directly probe how humans group and segment regions of natural images into objects. In this paradigm, two dots are placed on an image and participants are asked to indicate by button press whether they are on the same object or two different objects (Fig.~\ref{fig:beh_exp}A). One dot is always at the center of the image, and the other is at a peripheral location. The reaction time (RT) of this button press is the primary measure in this task and reveals the difficulty of object grouping. Critically, RT in this task has been shown to track known perceptual grouping effects such as the same-object advantage and Gestalt cues \cite{vecera2000toward, korjoukov2012time}, consistent with its use as a measure of object segmentation difficulty. Previous works using this paradigm have been limited in scale or have focused on simpler stimuli \cite{vecera2000toward, kim2019disentangling, korjoukov2012time}. For example, \cite{korjoukov2012time} used 24 hand-selected images depicting two instances of either a vehicle or an animal. Our work significantly scales up this effort.

\subsection{Behavioral Methods}

72 undergraduate students participated in our experiment for course credit. Their mean age was 20.4 years (range = 17–32) and all had normal or corrected-to-normal vision. This study was approved by the school Institutional Review Board. 

\paragraph{Stimuli and Apparatus:}
The images in our experiment were selected from the COCO2017 validation dataset \cite{lin2014microsoft}. Using the object-level annotations, we first selected images that had one object overlapping the center of the image, where we placed the center dot, ensuring that the reference point for the grouping judgment always fell within a well-defined object boundary. Among those, we selected images that had a second object overlapping the first. We then used an algorithm to place four peripheral markers (for same/different and close/far conditions) (Fig.~\ref{fig:beh_exp}B), ensuring that distances were matched between same and different conditions to prevent participants from making judgments based on distance alone. We only retained images where valid locations could be found for all four conditions. After this process, we manually inspected the remaining images and excluded those where dot placements fell outside object boundaries due to annotation inaccuracies or where the intended object was ambiguous upon visual inspection, leaving 288 images. Of these, 32 were used as practice trials and 256 as experimental trials. Fig.~\ref{fig:beh_exp}C shows the placement of dots across all four conditions. Each participant saw each image in only one condition (same-close, same-far, different-close, or different-far). We removed one experimental image from our analyses because the ground truth response was ambiguous due to the dot falling on a boundary shared by two objects, leaving 255 experimental images and 1020 (255$\times$4) trials for behavioral analyses and modeling experiments.


The assignment of images to the four conditions was counterbalanced across participants such that each image was seen in all four conditions across every group of four participants. The experiment was conducted on a 19-inch flat-screen CRT ViewSonic SVGA monitor with a screen resolution of 1024$\times$768 pixels and a refresh rate of 100 Hz. Participants were seated approximately 70 cm from the monitor, which subtended a visual angle of 30$^\circ$$\times$22$^\circ$. At this viewing distance, approximately 34 pixels spanned 1 degree of visual angle, placing the close peripheral dot approximately 3 degrees from the center dot and the far peripheral dot approximately 6 degrees from the center dot. Gaze position was recorded using an EyeLink 1000 eye-tracking system (SR Research) at a sampling rate of 1000 Hz. Fixations were parsed using the default EyeLink algorithm, with a velocity threshold of 30 degrees per second and an acceleration threshold of 8000 degrees per second squared. Calibration drift was checked before every trial, and recalibration was performed if necessary to ensure accurate eye-tracking.


\paragraph{Procedure:}
Participants were instructed to determine whether the two dots were on the same object or two different objects. Each trial started with the presentation of a central dot for 500 ms, indicating the location of the fixation point. Both the central and peripheral dots were then displayed for 1,000 ms without the image. Next, the dots were superimposed on the image and flickered at a frequency of 5 Hz to ensure their visibility. During the trial, participants were required to maintain their gaze on the center dot for the entire duration. If their gaze deviated more than 1 degree of visual angle from that location, the trial was terminated. 7\% of trials were excluded due to fixation breaks, defined as gaze deviating beyond this threshold. To record their responses, participants used a Microsoft gamepad controller, with buttons assigned to "same" or "different" responses. The button-hand assignment was randomized across participants to prevent dominant hand bias. Each participant performed 32 practice trials followed by 256 experimental trials, divided into four blocks of 64 trials each, with breaks provided between blocks. The order of image presentation within each block was randomized. Incorrect responses were indicated by an auditory feedback tone.

\begin{figure}[tb]
  \centering
  \includegraphics[height=8cm]{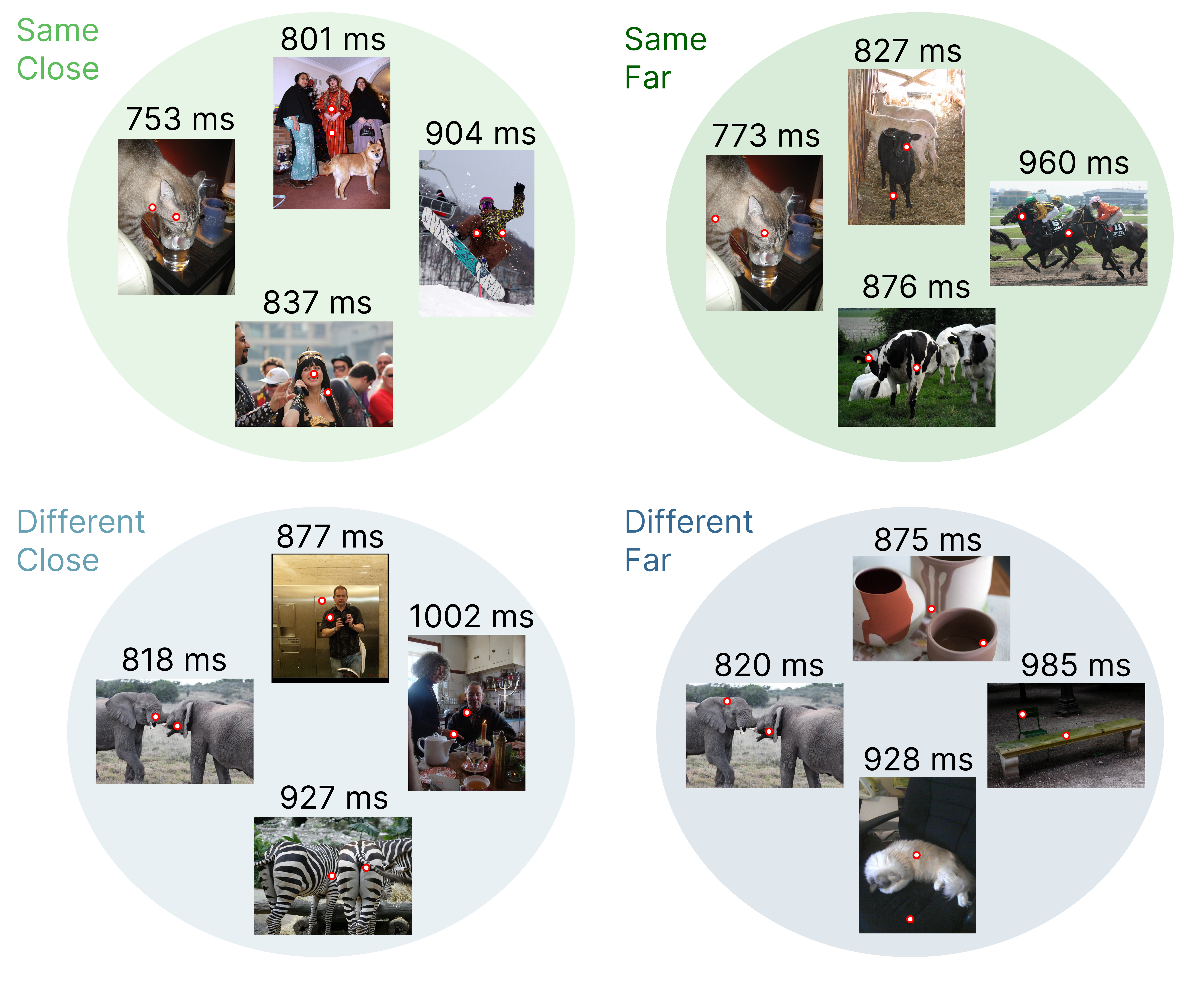} 
  \caption{
      Sample behavioral results for the four conditions in our experiment. The mean reaction time across subjects is displayed above.
  }
  \label{fig:sample_trials}
\end{figure}

\subsection{Behavioral Results}

The average subject accuracy on this task was 90\%. We report analyses only for trials where the subject response was correct; however, the patterns were largely the same when including both correct and incorrect trials. Fig.~\ref{fig:beh_exp}D shows the RT data for each condition. Subjects were faster to respond when the two dots were on the same object compared to when the peripheral dot was on a different object. This effect is known as the same-object advantage \cite{egly1994shifting}, indicating that the first dot facilitates selection of the whole object. This effect interacted with dot distance: we observed the fastest RTs in the close-separation same-object condition. When the peripheral dot was on a different object, dot separation had little effect on mean RTs, consistent with prior work \cite{roelfsema2023solving}. Detailed analyses of accuracy, sensitivity (d-prime), and response bias across subjects and conditions are provided in the supplementary.


Fig.~\ref{fig:sample_trials} shows four sample trials for each condition with comparable difficulties. For this visualization, we first ordered the trials within each condition by RT and selected the 50th, 100th, 150th, and 200th trials. Notably, RTs increase with dot distance when the dots are on the same object. Comparing the cat images (top row) illustrates this pattern. When the dots are on different objects, dot separation has little effect on RTs, as seen by comparing the elephant images (bottom row). While average and cross-condition behavioral patterns are informative, there is also interesting variability within each condition. Task difficulty, and RTs, increase when there are within-object boundaries between the dots, when the dots fall on different object parts with different textures, when they are on narrower parts of the object, when they are close to object boundaries, or when multiple objects from the same category are present. Our behavioral dataset therefore captures the variable conditions under which humans group objects in natural scenes. We will test models on how well they can predict the mean RT across subjects for each trial. 


\section{Modeling Experiments}
In our modeling experiments, we evaluate a diverse set of vision models, as shown in Table~\ref{tab:models}. We consider both Transformer and convolutional architectures. For Transformer models, we focus on the ViT-Base architecture \cite{dosovitskiy2020image}, including models trained with self-distillation (DINO \cite{caron2021emerging}, DINOv2 \cite{oquab2023dinov2, darcet2023vision}, and DINOv3 \cite{simeoni2025dinov3}), masked autoencoding (MAE \cite{he2022masked}), and supervised training on ImageNet-1K with or without ImageNet-21K pretraining \cite{steiner2021train} using the DeiT3 \cite{touvron2022deit} method. For these models, we extract patch features from the last Transformer layer. For convolutional models, we include ConvNext-Base \cite{liu2022convnet} models trained with supervised objectives on ImageNet-1K with or without ImageNet-21K pretraining, as well as a version distilled from a Transformer model trained with the DINOv3 objective. Input images are resized so that the resulting convolutional feature map dimensions match those of a ViT-Base model with patch size 16 on the original images. Features are then extracted from the last convolutional layer. The feature tensors (with size h $\times$ w $\times$ d) are then divided into h $\times$ w feature tokens of length d to represent different patches of the image. Further model implementation and training details are provided in the supplementary.

\begin{table}[tb]
  \caption{Vision models evaluated in this study and their grouping accuracy.}
  \label{tab:models}
  \centering
  \setlength{\tabcolsep}{6pt}
  \begin{tabularx}{\textwidth}{@{}
    l l l
    >{\raggedright\arraybackslash}X
    >{\raggedright\arraybackslash}X
  @{}}
    \toprule
    \makecell[tl]{Model} &
    \makecell[tl]{Architecture} &
    \makecell[tl]{Training \\regime} &
    \makecell[tl]{Large \\dataset} &
    \makecell[tl]{Grouping \\accuracy} \\
    \midrule
    DINOv3 ViT B      & Transformer   & Self-supervised & Yes & 91.9 \\
    DINOv2 ViT B      & Transformer   & Self-supervised & Yes & 89.0 \\
    DINO ViT B        & Transformer   & Self-supervised & No & 76.5 \\
    MAE ViT B         & Transformer   & Self-supervised & No  & 80.7 \\
    IN21k ViT B       & Transformer   & Supervised      & Yes & 72.2 \\
    IN1K ViT B        & Transformer   & Supervised      & No  & 70.6 \\
    IN21k ConvNext B  & Convolutional & Supervised      & Yes & 67.4 \\
    IN1K ConvNext B   & Convolutional & Supervised      & No  & 60.0 \\
    DINOv3 ConvNext B & Convolutional & Distilled       & Yes & 86.7 \\
    \bottomrule
  \end{tabularx}
\end{table}

To predict human grouping judgments, we extract patch features from the two dot locations in each image and concatenate them to form a trial representation, with the central patch feature concatenated first followed by the peripheral patch feature. A two-layer MLP readout is then trained to predict the same or different response from this concatenated representation. To provide sufficient training data for this readout, we applied the same dot placement algorithm to the COCO2017 training set to generate approximately 30,000 trials, which were used to train the MLP classifier. The model was then evaluated on the 1,020 held-out behavioral trials.
Grouping accuracy results are shown in Table~\ref{tab:models}. Self-supervised Transformer models trained with the DINO objective achieve the strongest performance, with DINOv3 ViT B reaching 91.9\% accuracy. Supervised Transformer models perform considerably lower despite sharing the same architecture, suggesting that the training objective rather than architecture alone drives object-centric representations. MAE ViT B (80.7\%) falls between the self-supervised DINO models and the supervised models, consistent with it learning some object-centric structure but not to the same degree as self-distillation. Among convolutional models, DINOv3 ConvNext B (86.7\%) substantially outperforms its supervised counterparts, further highlighting the quality of the DINOv3 dense features.

\subsection{Behavior Prediction}

To predict trial-by-trial reaction times, we use the same 2-layer MLP readout procedure but train it on the 1,020 behavioral trials using nested cross-validation. In the outer loop, we use 10-fold cross-validation, training on 90\% of the trials and testing on the held-out 10\%. Within each outer fold, the 90\% training data is further split into a 90/10 train-validation split used for early stopping and hyperparameter selection. We train 10 random seeds per outer fold and average their predictions, resulting in a robust estimate of each model's ability to predict human RTs. To account for noise in the behavioral measurements, we normalize model performance by a human noise ceiling. This ceiling is estimated independently of any model: we randomly split the subjects into two equal halves 20 times, compute the Spearman correlation between the mean RTs of the two halves for each split, and average across the 20 splits to obtain a single noise ceiling value. Separately, for each model we compute the Spearman correlation between the model's predicted RTs and the mean RTs of each of the 20 subject splits, and average these correlations across splits to obtain the model-human correlation. The normalized Spearman correlation is then obtained by dividing this model-human correlation by the noise ceiling, such that a value of 1.0 would indicate alignment with human behavior at the level of the noise ceiling. Raw (non-normalized) Spearman correlations are reported in the supplementary.

\begin{figure}[tb]
  \centering
  \includegraphics[width=0.99\textwidth]{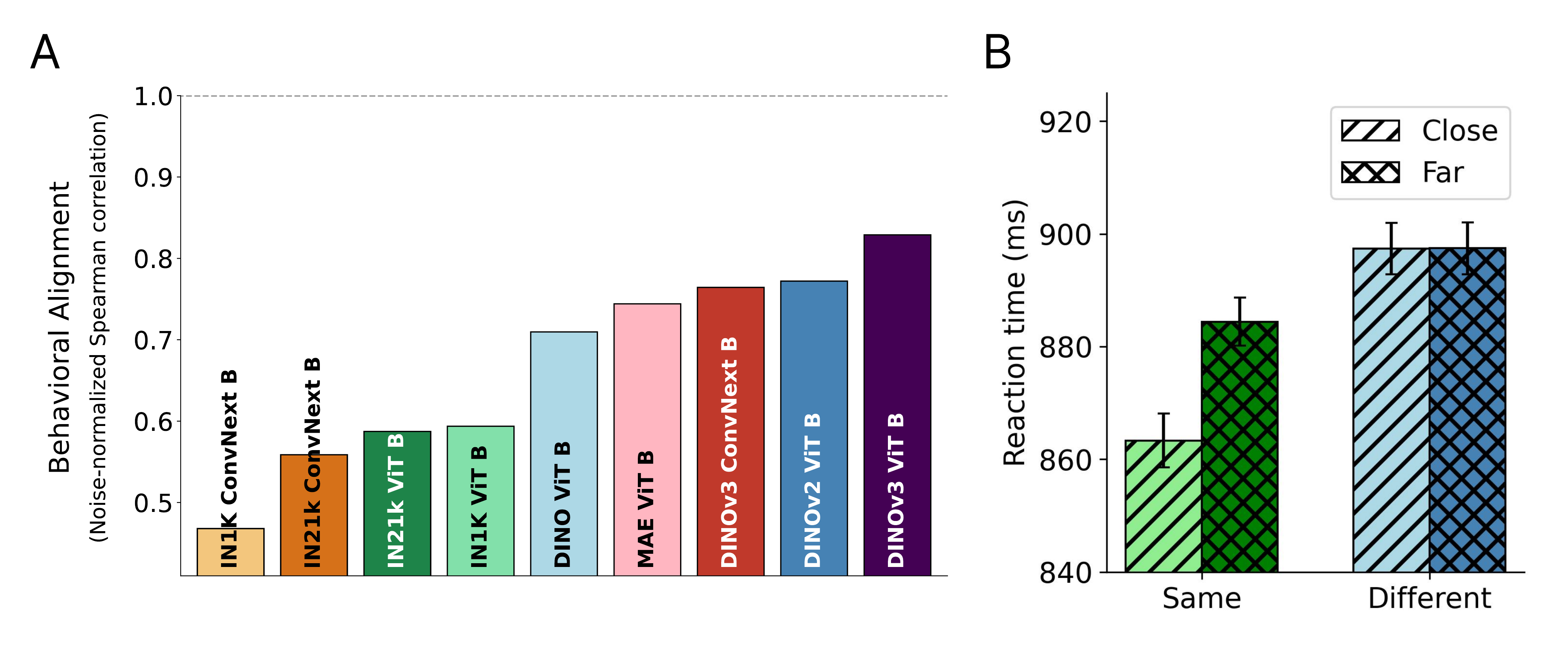} 
  \caption{
      \textbf{A)} Noise-normalized Spearman correlation between model-predicted and human reaction times across all models, ordered from lowest to highest. Models trained with self-supervised DINO objectives consistently outperform supervised counterparts with the same architecture. \textbf{B)} Mean reaction times predicted by DINOv3 ViT B for each experimental condition. The model reproduces the key signatures of human grouping behavior, including faster responses for same-object trials and a distance effect that is specific to the same-object condition.
  }
  \label{fig:model_pred}
\end{figure}

Results are shown in Fig.~\ref{fig:model_pred}A. The same ordering observed in the grouping task largely holds here: self-supervised Transformer models trained with the DINO objective achieve the strongest behavioral alignment, with DINOv3 ViT B reaching the highest normalized Spearman correlation among all models tested. Supervised models show considerably weaker alignment despite sharing the same architecture, again pointing to the training objective as the key factor. MAE ViT B falls in an intermediate range, and convolutional models show the weakest alignment overall, with DINOv3 ConvNext B being a notable exception. Fig.~\ref{fig:model_pred}B shows the mean RTs predicted by the best-performing model, DINOv3 ViT B, broken down by condition. The predicted RTs show the same-object advantage and the distance effect within the same-object condition observed in the behavioral data, confirming that the model captures the qualitative structure of human object grouping behavior.


\subsection{Object-Centric Representations}

\begin{figure}[t]
  \begin{center}
    \centerline{\includegraphics[width=0.90\textwidth]{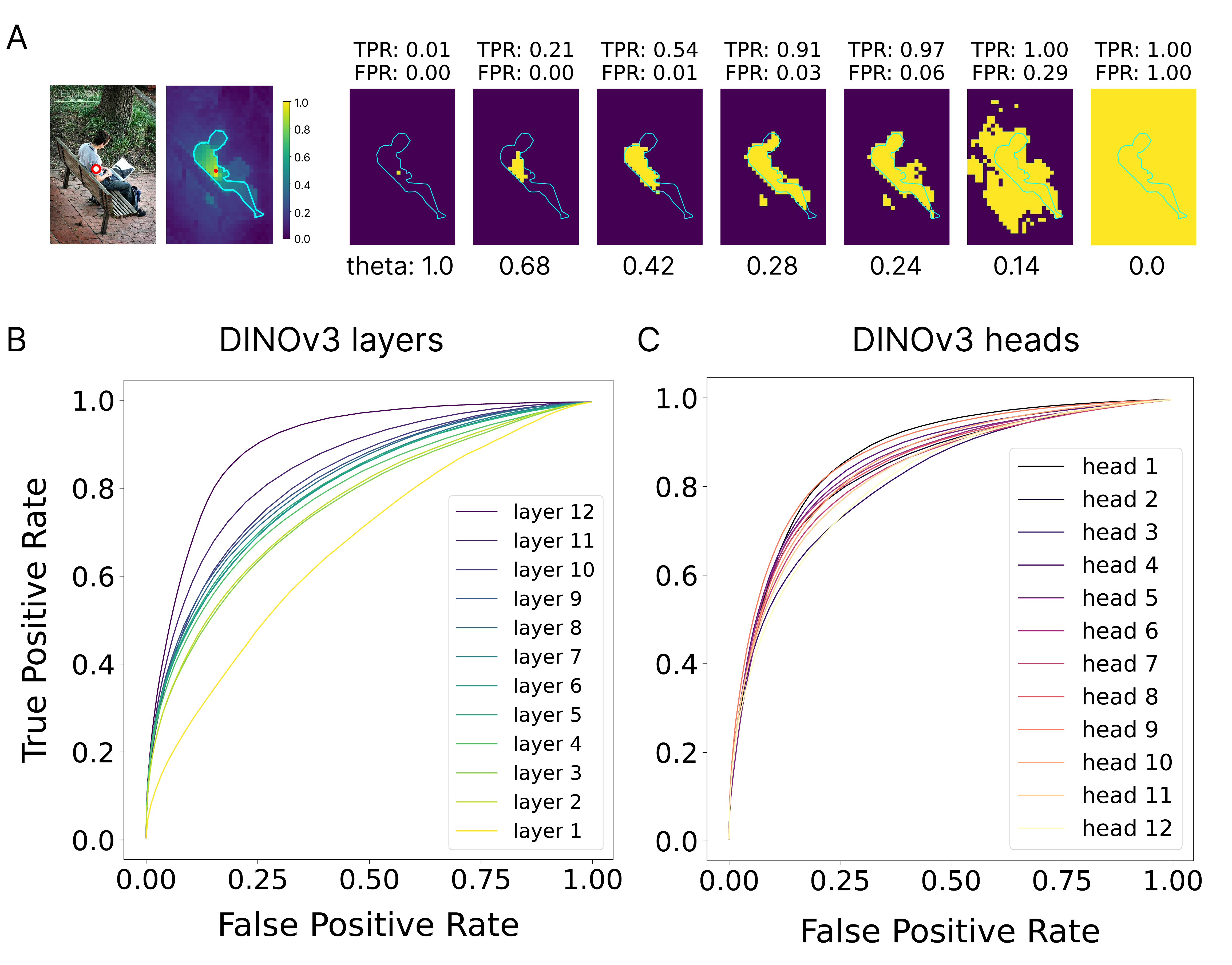}}
    \caption{
      \textbf{A)} A sample experimental trial with the central dot shown on the top left, alongside its affinity map (using DINOv3 features) showing normalized feature similarity between the central patch and all other patches (colorbar shown). For decreasing threshold values ($\theta$), patches with affinity above the threshold are shown in yellow. The TPR and FPR are displayed above each thresholded map. TPR increases substantially before FPR rises, indicating a strong object-centric signal in the affinity map. \textbf{B)} ROC curves averaged across all trials for each of the 12 layers of the DINOv3 ViT B model. The legend is ordered by decreasing AUC, with deeper layers showing stronger object-centric structure. \textbf{C)} ROC curves for features from different attention heads of the last layer of DINOv3 ViT B, showing broadly similar object-centricity across heads.
      }
    \label{fig:object_centric}
  \end{center}
\end{figure}

To quantify the object-centric structure in model representations, we compute affinity maps by calculating the cosine similarity of each patch token's feature with all other tokens, yielding a measure of feature similarity between each patch and the rest of the image. Sample affinity maps for a few patch locations are shown in Fig.~\ref{fig:affinity_maps}A, generated using patch features from the DINOv3 ViT B model. We then perform an ROC analysis across our experimental images to quantify how well these affinity signals align with ground truth object boundaries. For each trial, we compute the affinity map from the central dot location (one sample map, also from the DINOv3 ViT B model and taken from Fig.~\ref{fig:affinity_maps}A, is shown on the left of Fig.~\ref{fig:object_centric}A) and apply a range of thresholds to assess the spatial distribution of active patches relative to the ground truth object boundary. The True Positive Rate (TPR) is computed as the proportion of within-object patches whose affinity exceeded the threshold, and the False Positive Rate (FPR) as the proportion of outside-object patches whose affinity exceeded the threshold. As shown in Fig.~\ref{fig:object_centric}A, in this case TPR increases substantially as the threshold decreases while FPR remains low, indicating that the patches most similar to the central patch tend to be on the same object. Only once TPR reaches a high level does FPR begin to rise, reflecting a strong object-centric signal in the affinity map.
Averaging TPR and FPR across all trials yields a summary ROC curve for each model. Fig.~\ref{fig:object_centric}B shows ROC curves for all 12 layers of DINOv3 ViT B, with deeper layers showing progressively stronger object-centric structure (legend ordered by decreasing AUC). Fig.~\ref{fig:object_centric}C shows ROC curves across the 12 attention heads of the last layer, showing that object-centric structure is broadly distributed across heads, with each head contributing similarly to patch-level object representation. 


\begin{figure}[tb]
  \centering
  \includegraphics[width=0.99\textwidth]{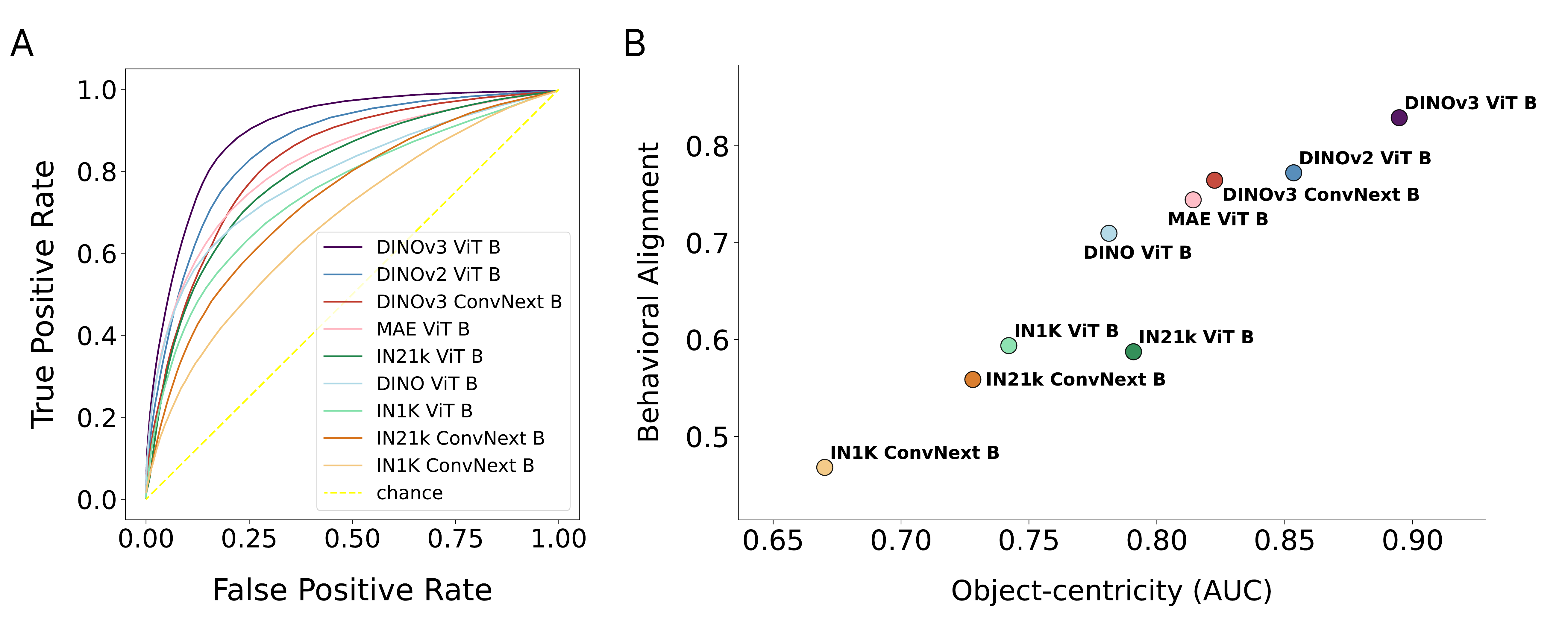} 
  \caption{
    \textbf{A)} ROC curves quantifying the object-centricity of patch-level representations for all models evaluated in this study, using features from the final layer. The legend is sorted by decreasing AUC, with self-supervised DINO-based models consistently achieving higher object-centricity than supervised or reconstruction-based counterparts. The diagonal dashed line indicates chance performance. \textbf{B)} Scatter plot relating each model's object-centricity (AUC) to its noise-normalized Spearman correlation with human reaction times. Each circle represents one model. Models with stronger object-centricity tend to exhibit greater alignment with human perceptual behavior, with the relationship holding across both Transformer and convolutional architectures. The correlation between object-centric AUC and behavioral alignment across all 9 models is Spearman r=0.950, p=0.0001.}
  \label{fig:auc_rt}
\end{figure}

To compare object-centricity across all models, we compute the AUC of the ROC curve for each model using features from its last layer. Fig.~\ref{fig:auc_rt}A shows the ROC curves for all models, ordered by decreasing AUC in the legend. Self-supervised Transformer models trained with the DINO objective achieve the highest AUC values, reflecting strong object-centric structure in their patch representations. DINOv3 ViT B leads across all models, followed closely by DINOv2 ViT B, while supervised Transformer and convolutional models show progressively weaker object-centricity. Notably, DINOv3 ConvNext B, despite being a convolutional model, achieves a substantially higher AUC than its supervised convolutional counterparts, suggesting that strongly object-centric representations, learned through distillation, can be computed in this architecture as well. MAE ViT B falls in an intermediate range, consistent with it learning some degree of object-centric structure through reconstruction but not to the same extent as self-distillation models.
Fig.~\ref{fig:auc_rt}B shows the relationship between each model's object-centric AUC and its behavioral alignment (noise-normalized Spearman correlation). The two measures are strongly correlated across models: models with stronger object-centric structure in their representations also predict human reaction times more accurately. This relationship holds across both architecture types and training objectives, suggesting that object-centricity is a general principle linking model representations to human perceptual behavior. Together, these results indicate that the degree to which a model encodes object identity at the patch level is a key factor driving its alignment with human object grouping, and that the DINO family of training objectives promotes this property more effectively than supervised or reconstruction-based alternatives. Scatter plots showing correlations between best model predictions and the behavioral RTs for each of the four conditions are provided in the supplementary results section.

\subsection{Gram Alignment}

\begin{figure}[tb]
  \centering
  \includegraphics[width=0.95\textwidth]{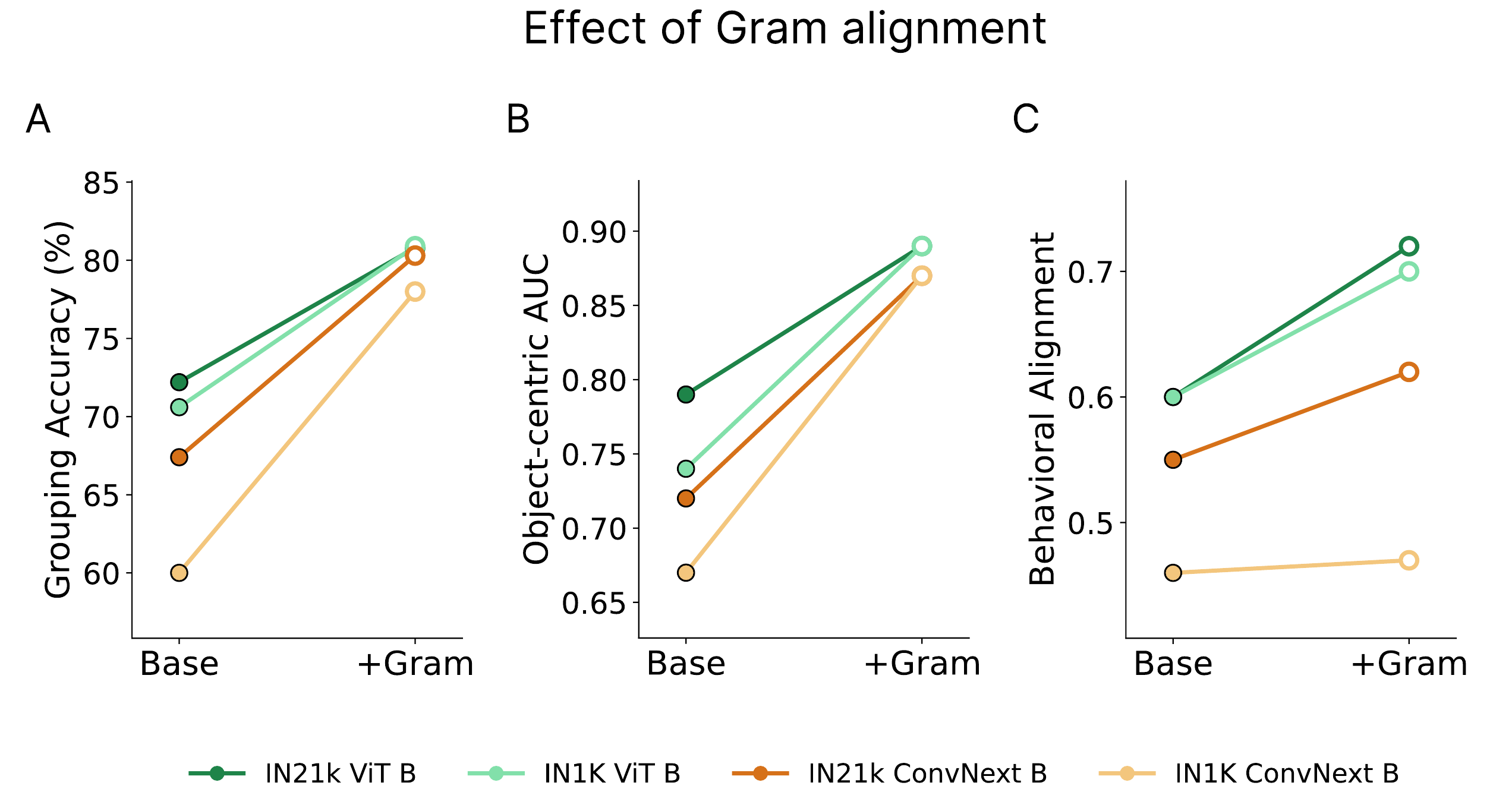} 
  \caption{
  Effect of Gram matrix alignment on model performance across three metrics. Each panel shows the change from the base supervised model (left) to the Gram-aligned version (right) for four models: IN21k ViT B, IN1K ViT B, IN21k ConvNext B, and IN1K ConvNext B. Gram alignment is performed by fine-tuning each model on ImageNet classification while distilling the Gram matrix structure from DINOv3 ViT B. \textbf{A)} Grouping accuracy improves substantially for all models following alignment. \textbf{B)} Object-centric AUC increases consistently, with aligned models approaching the performance of self-supervised DINO models. \textbf{C)} Behavioral alignment with human reaction times improves across all models, with Transformer-based models showing larger gains than convolutional models.
  }
  \label{fig:gram_effect}
\end{figure}

The Gram matrix captures the pairwise similarity structure across all feature vectors in a representation. Here, for a model with $h\times w$ patch tokens, we compute the cosine similarity between every pair of patch feature vectors, yielding an $(h\times w) \times (h\times w)$ matrix where each entry reflects the degree to which two patches share similar features. Fig.~\ref{fig:affinity_maps}B shows an example Gram matrix, where brighter values indicate stronger feature similarity between patch pairs. Importantly, in models with strong object-centric representations, patches belonging to the same object tend to have higher pairwise similarities, resulting in a block-like structure in the Gram matrix that reflects object boundaries.

To investigate whether explicitly aligning the Gram matrix of supervised models with that of a self-supervised model improves their representations, we fine-tune four supervised models, IN21k ViT B, IN1K ViT B, IN21k ConvNext B, and IN1K ConvNext B, on ImageNet classification while simultaneously distilling the Gram matrix structure from DINOv3 ViT B (Fig.~\ref{fig:gram_effect}). Gram alignment consistently improves grouping accuracy across all four models (Fig.~\ref{fig:gram_effect}A), with gains ranging from approximately 8 to 18 percentage points. A similarly consistent improvement is observed in object-centric AUC (Fig.~\ref{fig:gram_effect}B), with all models showing substantially higher object-centricity after Gram alignment, approaching the AUC values of the self-supervised DINO models. Behavioral alignment also improves across all models (Fig.~\ref{fig:gram_effect}C), though the magnitude of improvement varies. Notably, the two Transformer models show larger gains than the ConvNext models across all three metrics, suggesting that the Transformer architecture is more amenable to Gram matrix alignment, likely due to its self-attention mechanism already encoding pairwise patch relationships. Interestingly, the gains from Gram alignment are substantial even for models without large-scale pretraining, suggesting that Gram alignment can partially compensate for the absence of large dataset training, particularly in Transformer models where the gains are most pronounced. Taken together, these results demonstrate that matching the Gram matrix structure of a self-supervised model can substantially improve the object-centricity and behavioral alignment of supervised models, pointing to feature correlation structure across image patches as a key mechanism underlying human-like object representations. A CLS token distillation baseline isolating Gram structure as the active ingredient is reported in the supplementary results section.

\section{Conclusion}

Classical theories of perceptual grouping have long emphasized the role of feature similarity in binding image regions into coherent objects, from Gestalt principles to association fields and lateral connectivity in visual cortex \cite{field1993contour, roelfsema2023solving}. Our results show that self-supervised vision Transformers implicitly instantiate a similar computational principle: patches belonging to the same object develop highly similar feature representations, and the strength of this object-centric structure predicts how readily humans group those regions. The same-object advantage and distance effects we observe in human reaction times are captured by these models, suggesting that the patch-level similarity structure learned through self-supervised training reflects key aspects of how the visual system organizes scenes into objects. This emergent alignment between self-supervised training objectives and human object perception supports the view that these objectives better approximate the organizational principles of perception than supervised classification. A gap nonetheless remains between the best-performing model and the human noise ceiling, leaving room for future models to close this distance.

Our findings also offer guidance on model design. The consistent advantage of DINO-trained models over supervised counterparts with identical architectures demonstrates that the training objective is a major factor beyond architecture alone for object-centric representations. At the same time, the strong performance of DINOv3 ConvNext B relative to its supervised convolutional counterparts shows that convolutional architectures can also instantiate similar representations. However, the greater responsiveness of Transformer models to Gram matrix alignment suggests that the self-attention mechanism provides a particularly natural substrate for encoding pairwise patch relationships. It is worth noting that perfect object-centricity is not necessarily the goal: representations entirely organized around object boundaries would resemble semantic segmentation maps, losing the fine-grained feature structure that supports other visual tasks. The most behaviorally aligned models may therefore be those that strike an optimal balance between object-level organization and feature sensitivity.

The effect of Gram matrix alignment offers the most mechanistically informative finding of our work. By fine-tuning supervised models to match the pairwise feature correlation structure of DINOv3 ViT B while maintaining ImageNet classification performance, we demonstrate that Gram matrix structure contributes to object-centricity and behavioral alignment. This finding converges with the design of DINOv3 itself, where Gram matrix anchoring was introduced as an explicit training signal to preserve and enhance the dense feature quality of the model \cite{simeoni2025dinov3}. These results show that feature correlation across patches plays an important role in human-like visual representations. Together, these findings suggest that the path toward more perceptually aligned vision models lies not in scaling classification objectives, but in explicitly shaping the similarity structure of learned representations to reflect the object-level organization of natural scenes.

\section*{Acknowledgements}
Research reported in this publication was supported in part by the National Institute of Neurological Disorders and Stroke of the National Institutes of Health under award numbers 1RF1NS128897 and 4R01NS128897. This work was also supported by Hankuk University of Foreign Studies Research Fund to SA. The contribution of M.Z. was supported by the National Research Foundation, Singapore under its NRFF award NRF-NRFF15-2023-0001 and a Startup Grant from Nanyang Technological University, Singapore. Additionally, we would like to thank the National Science Foundation for supporting this work through awards 2123920 and 2444540, and the National Institutes of Health through their award R01EY030669, to GZ. The content is solely the responsibility of the authors and does not necessarily reflect the views of the funding agencies.

\newpage

%
%
\bibliographystyle{splncs04}
\bibliography{main}

\newpage

\section{Supplementary Material}

\subsection{Behavioral Analyses}

\subsubsection{Subject-level Accuracy, Sensitivity, and Criterion}

Accuracy is consistent across subjects (mean=0.90, SD=0.05, no subject below 0.78), confirming that the task was well understood and performed reliably. d-prime is high across subjects (mean=2.78, SD=0.65), indicating strong perceptual sensitivity to the same/different distinction. The criterion is tightly centered around zero (mean=$-$0.03, SD=0.17), confirming no systematic response bias.

\subsubsection{Accuracy by Condition}

Accuracy is consistently high across all conditions (same-close: 0.92, same-far: 0.90, diff-close: 0.89, diff-far: 0.91), with no condition showing floor or ceiling effects. The small variation across conditions is consistent with the same-object advantage: same-close is the easiest condition and diff-close is the hardest, as close dots on different objects are the most confusable. 

\subsubsection{d-prime and Criterion by Distance Condition}

d-prime is constant across close and far conditions (close: 2.79$\pm$0.67, far: 2.84$\pm$0.73), confirming equal sensitivity to the same/different distinction regardless of dot separation. The criterion is near zero in both conditions (close: $-$0.10$\pm$0.21, far: 0.05$\pm$0.20), indicating no systematic response bias in either condition. Together these results confirm that the speed-accuracy tradeoff did not differentially affect performance across conditions.

\subsubsection{Reaction Time Distributions}

As a measure of the overlap between same and different RT distributions, we computed the AUC of an ROC analysis classifying same vs. different object trials from trial-averaged RTs ($\approx$18 subjects per trial). The AUC is substantially above chance in both close (AUC=0.71, p$<$0.0001) and far (AUC=0.62, p$<$0.0001) conditions, confirming that the behavioral signal used for model comparison is well above chance. Single-trial AUC values are lower (close: AUC=0.59, far: AUC=0.55), consistent with well-known trial-to-trial RT variability. The same-object advantage is isotropic across all eight directions (std of mean RT: 14.7 ms for same, 24.6 ms for different).

\subsection{Model Implementation Details}

\subsubsection{MLP Readout Architecture and Training}

For both the grouping and RT prediction tasks, the MLP readout takes as input the concatenation of two patch feature vectors, one from the central dot location and one from the peripheral dot location. The input dimension is therefore $2\times768$ for Transformer models and $2\times1024$ for ConvNext models. The MLP consists of two linear layers with ReLU activation, with hidden dimension equal to the feature dimension, followed by a final output layer.

For the grouping task, the output layer produces a two-class prediction (same or different object), trained with CrossEntropy loss. Training uses the Adam optimizer with learning rate $5\times10^{-4}$, a learning rate drop at epoch 15, and 30 epochs with gradient accumulation over 32 steps.

For the RT prediction task, the output layer produces a scalar prediction of the z-scored mean reaction time, where targets are normalized using the mean and standard deviation computed from the training set only. Training is performed with SmoothL1 loss using the Adam optimizer with learning rate $1\times10^{-5}$, a learning rate drop at epoch 40, and 80 epochs with gradient accumulation over 8 steps.

\subsubsection{Gram Alignment Training}

The Gram-aligned models are produced by fine-tuning only the last block of each supervised model on ImageNet classification while simultaneously distilling the Gram matrix structure from DINOv3 ViT B. The training objective combines two losses: a standard CrossEntropy classification loss (weight 1) and a Gram MSE loss (weight 10), where the Gram loss minimizes the MSE between the pairwise cosine similarity matrix of the student's last block features and those of the DINOv3 ViT B teacher. The higher weight on the Gram loss is chosen to make the two loss values comparable in scale during training. Training uses the Adam optimizer with learning rate $5\times10^{-4}$.

The Gram loss results in no change in ImageNet top-1 accuracy relative to fine-tuned baselines trained with the same procedure but without the Gram loss (IN1K ViT B: 0.77, IN21k ViT B: 0.78, IN1K ConvNext B: 0.74, IN21k ConvNext B: 0.76), confirming that improvements in object-centricity and behavioral alignment are not a consequence of altered classification performance.

\subsection{Supplementary Results}

\subsubsection{Linear Readout Baseline}

To verify that behavioral alignment results are not an artifact of MLP capacity, we trained a linear readout for RT prediction across all 9 models using the same nested cross-validation procedure. The model ranking is fully preserved and the correlation between object-centric AUC and behavioral alignment remains high (Spearman $r$=0.917, $p$=0.0005), confirming that backbone object-centricity rather than readout capacity drives the result.

\subsubsection{CLS Token Distillation Baseline}

To isolate Gram structure as the active ingredient in the Gram alignment experiments, we fine-tuned IN21k ViT B and IN1K ViT B using CLS token distillation from the same DINOv3 teacher instead of Gram distillation. The CLS loss minimizes $1 - \cos(\mathbf{z}_s, \mathbf{z}_t)$, where $\mathbf{z}_s$ and $\mathbf{z}_t$ are the student and teacher CLS token embeddings, driving the student's global representation to align with the teacher's without imposing any patch-level correlation structure. CLS distillation shows no improvement in behavioral alignment over the base models (IN21k ViT B: 0.59$\rightarrow$0.59, IN1K ViT B: 0.59$\rightarrow$0.58), while Gram distillation shows substantial improvement (0.59$\rightarrow$0.72, 0.59$\rightarrow$0.70), directly isolating patch-level correlation structure as the active ingredient.

\begin{figure}[tb]
\centering
\includegraphics[width=0.99\textwidth]{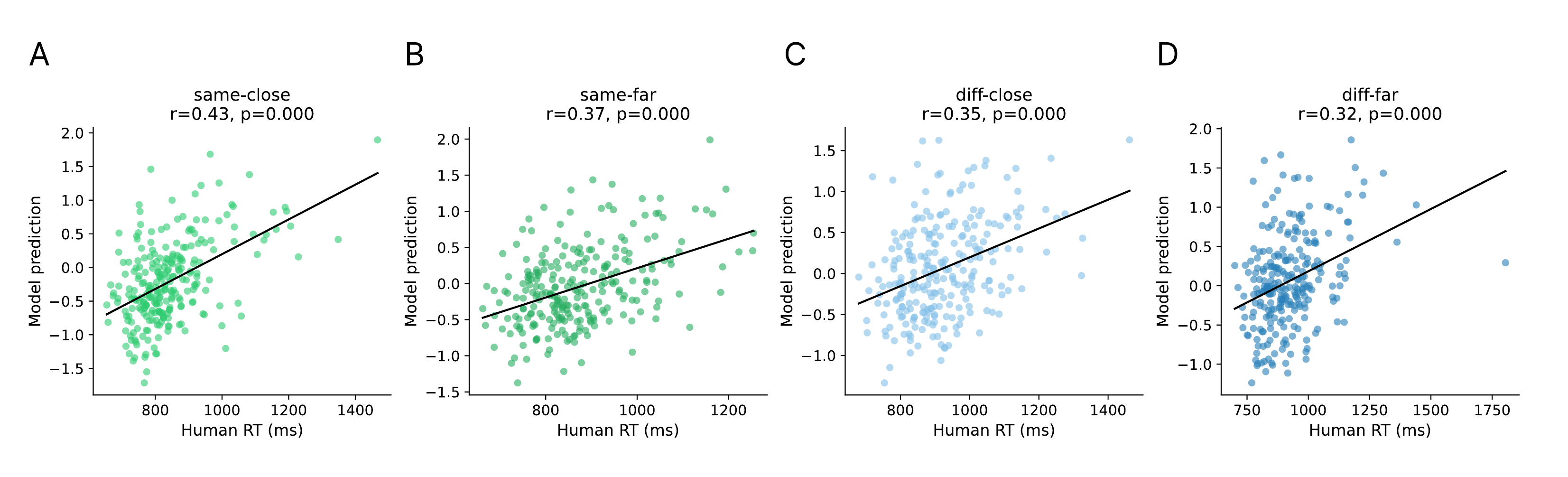}
\caption{Scatter plots of model-predicted vs. human mean RTs for the best-performing model (DINOv3 ViT B), broken down by condition. Model predictions are in z-scored units (normalized using training-set statistics only); human RT is in milliseconds. Each point represents one trial (255 per condition). The positive relationship between model predictions and human RTs is significant in all four conditions independently (same-close: $r$=0.43, same-far: $r$=0.37, diff-close: $r$=0.35, diff-far: $r$=0.32, all p$<$0.0001), confirming that behavioral alignment is not driven by a single condition, near-constant predictions, or outliers.}
\label{fig:supp_scatter_condition}
\end{figure}

\subsubsection{Model-Human Condition-Split Correlations}

Figure~\ref{fig:supp_scatter_condition} shows scatter plots of model-predicted vs. human mean RTs for each of the four conditions separately, for the best-performing model (DINOv3 ViT B). Model-human correlations are significant in all four conditions (same-close: $r$=0.43, same-far: $r$=0.37, diff-close: $r$=0.35, diff-far: $r$=0.32, all p$<$0.0001), confirming that the behavioral alignment is not driven by a single condition, near-constant predictions, or a small number of outliers. Table~\ref{tab:supp_condition_corr} reports condition-split correlations for both the best and worst performing models.

\begin{table}[tb]
\centering
\caption{Spearman $r$ between model-predicted and human mean RTs by condition. All p$<$0.01.}
\setlength{\tabcolsep}{4pt}
\small
\begin{tabular}{lcc}
\toprule
Condition & DINOv3 ViT B & IN1K ConvNext B \\
\midrule
Same-close & 0.43 & 0.33 \\
Same-far   & 0.37 & 0.28 \\
Diff-close & 0.35 & 0.20 \\
Diff-far   & 0.32 & 0.20 \\
\bottomrule
\end{tabular}
\label{tab:supp_condition_corr}
\end{table}

\subsubsection{Raw and Normalized Spearman Correlations}

Table~\ref{tab:supp_raw_corr} reports the raw and noise-ceiling-normalized Spearman correlations between model-predicted and human mean RTs for all 9 models. The noise ceiling is estimated by splitting subjects into two equal halves 20 times and averaging the Spearman correlation between the two half-means across splits (mean=0.42, std=0.037). For each model, the Spearman correlation between model predictions and each of the 20 subject split means is computed, averaged across splits, and divided by the noise ceiling once as a final normalization step.

\begin{table}[h]
\centering
\caption{Raw and normalized Spearman $r$ for all models. Noise ceiling = 0.42.}
\setlength{\tabcolsep}{4pt}
\small
\begin{tabular}{lcc}
\toprule
Model & Raw $r$ & Norm. $r$ \\
\midrule
DINOv3 ViT B      & 0.348 & 0.83 \\
DINOv2 ViT B      & 0.324 & 0.77 \\
DINOv3 ConvNext B & 0.321 & 0.76 \\
MAE ViT B         & 0.313 & 0.74 \\
DINO ViT B        & 0.298 & 0.71 \\
IN21k ViT B       & 0.247 & 0.59 \\
IN1K ViT B        & 0.249 & 0.59 \\
IN21k ConvNext B  & 0.235 & 0.56 \\
IN1K ConvNext B   & 0.197 & 0.47 \\
\bottomrule
\end{tabular}
\label{tab:supp_raw_corr}
\end{table}

\end{document}